\documentclass{article}

\usepackage[final]{nips_2016}

\usepackage[utf8]{inputenc} 
\usepackage[T1]{fontenc}    
\usepackage{url}            
\usepackage{booktabs}       
\usepackage{amsfonts}       
\usepackage{nicefrac}       
\usepackage{microtype}      

\usepackage{tikz}
\usepackage{pgfplots}
\usepackage{amsmath}
\usepackage{subfig} 
\usepackage{mathtools}
\usepackage{lipsum}

\title{Quantized neural network design under \\ weight capacity constraint}

\author{
 Sungho Shin, Kyuyeon Hwang, and Wonyong Sung\\
 Department of Electrical and Computer Engineering\\
 Seoul National University\\
  Seoul, 08826 Korea \\
  \texttt{sungho.develop@gmail.com, kyuyeon.hwang@gmail.com, wysung@snu.ac.kr} \\
}  

\begin{document}

\maketitle

\begin{abstract}
The complexity of deep neural network algorithms for hardware implementation can be lowered either by scaling the number of units or reducing the word-length of weights. Both approaches, however, can accompany the performance degradation although many types of research are conducted to relieve this problem. Thus, it is an important question which one, between the network size scaling and the weight quantization, is more effective for hardware optimization. For this study, the performances of fully-connected deep neural networks (FCDNNs) and convolutional neural networks (CNNs) are evaluated while changing the network complexity and the word-length of weights. Based on these experiments, we present the \emph{effective compression ratio} (ECR) to guide the trade-off between the network size and the precision of weights when the hardware resource is limited.
\end{abstract}

\section{Introduction}
\label{Introduction}
Deep neural networks (DNNs) begin to find many real-time applications, such as speech recognition, autonomous driving, gesture recognition, and robotic control \citep{sak2015fast,chen2015deepdriving,shin2016dynamic,corradini2015robust}. Recent works show that the precision required for implementing fully-connected deep neural networks (FCDNNs), convolutional neural networks (CNNs) or recurrent neural networks (RNNs) needs not be very high, especially when the quantized networks are trained again to learn the effects of lowered precision. In the fixed-point optimization examples shown in \citet{hwang2014fixed}, neural networks with ternary weights showed quite good performance which was close to that of floating-point arithmetic. However, the performance of DNNs usually degrades when the weights are represented using a very low precision. Thus, we have a question whether it might be a better option to reduce the network size, instead of severely quantizing the weights, for efficient implementations.

In this work, we compare the performance of FCDNNs and CNNs under two constraints for hardware implementation, one is reducing the network size and the other is lowering the precision of the weights. We conduct the experiments with FCDNNs for phoneme recognition and CNNs for image classification. To control the network complexity, the number of units in each hidden layer is varied in the FCDNN. For the CNN, the number of feature maps for each layer is changed. The retraining based quantization algorithm is used for fixed-point optimization of weights \citep{hwang2014fixed}.

Based on the experiments, we propose a metric called the \emph{effective compression ratio (ECR)} that compares the complexity of floating-point and fixed-point networks showing the same performance. This analysis intends to provide a guideline to network size and word-length determination for efficient hardware implementation of deep neural networks (DNN).  

\section{Related Work}
\label{sec_related}
Fixed-point design of DNNs with ternary weights show quite good performances that are very close to the floating-point results \citep{hwang2014fixed,anwar2015fixed,shin2016fixed}. 
The ternary weight based FCDNN is used for VLSI implementations, by which the algorithms can operate with only on-chip memory consuming very low power \citep{kim2014x1000}. The CNN is implemented by XNOR-bitcounting operations~\citep{rastegari2016xnor}. Binary weight based deep neural network design is also studied \citep{courbariaux2015binaryconnect}. Pruned floating-point weights are utilized for efficient GPU-based implementations, where small valued or less important weights are forced to zero to reduce the number of arithmetic operations and the memory space for weight storage \citep{yu2012exploiting,han2015deep,anwar2015structured}. 

Most of the above works are experimented using large size neural networks. However, mobile or embedded portable devices have limited resources, and thus small size fixed-point networks showing good performances are very needed.
  
\section{Fixed-Point FCDNN and CNN Design}
\label{sec_fixedpoint}

This section explains the design of FCDNN and CNN with varying network complexity and weight precision.  

\subsection{FCDNN and CNN Design}

In this work, we examine an FCDNN for phoneme recognition and a CNN for image classification. The reference DNN has four hidden layers. Each of the hidden layers has $N_{h}$ units; the value of $N_{h}$ is altered to control the complexity of the network. 
We conduct experiments with $N_{h}$ value of 32, 64, 128, 256, 512, and 1024. The input layer of the network has 1,353 units to accept 11 frames of a Fourier-transform-based filter-bank with 40 coefficients ($+$energy) distributed on a mel-scale, together with their first and second temporal derivatives. The output layer consists of 61 softmax units which correspond to 61 target phoneme labels \citep{mohamed2012acoustic}. Phoneme recognition experiments were performed on the TIMIT corpus. 

The CNN used is for CIFAR-10 dataset~\citep{krizhevsky2009learning}. It contains a training set of 50,000 and a test set of 10,000 images. We divided the training set to 40,000 images for training and 10,000 images for validation. The reference CNN has 3 convolution and max-pooling layers, a fully connected hidden layer with 64 units, and the output with 10 softmax units. We control the number of feature maps in each convolution layer. The reference size has 32-32-64 feature maps with a 5 by 5 kernel size as used in~\citet{krizhevskey2014cuda}. To know the effects of network size variation, the number of feature maps is reduced or increased. 
The configurations of the feature maps used for the experiments are 8-8-16, 16-16-32, 32-32-64, 64-64-128, 96-96-192, and 128-128-256. Note that the fully connected layer in the CNN is not changed. 

\subsection{Fixed-Point Optimization of DNNs}

Reducing the word-length of weights brings several advantages in hardware based implementation of neural networks.
First, it lowers the arithmetic precision, and thereby reduces the number of gates needed for multipliers.  
Second, the size of memory for storing weights is minimized, which would be a big advantage when keeping them on a chip, instead of external DRAM or NAND flash memory.
Note that FCDNNs demand a very large number of weights.
Third, reduced arithmetic precision or minimization of off-chip memory accesses leads to low power consumption.

The fixed-point DNN algorithm design consists of three steps: floating-point training, direct quantization, and retraining of weights. Refer to~\citet{hwang2014fixed} for the details.
\section{Analysis of Quantization Effects}
\label{sec:experiments}
\begin{figure}[h]
\centering
\subfloat[FCDNN]{
\includegraphics[width=0.45\linewidth]{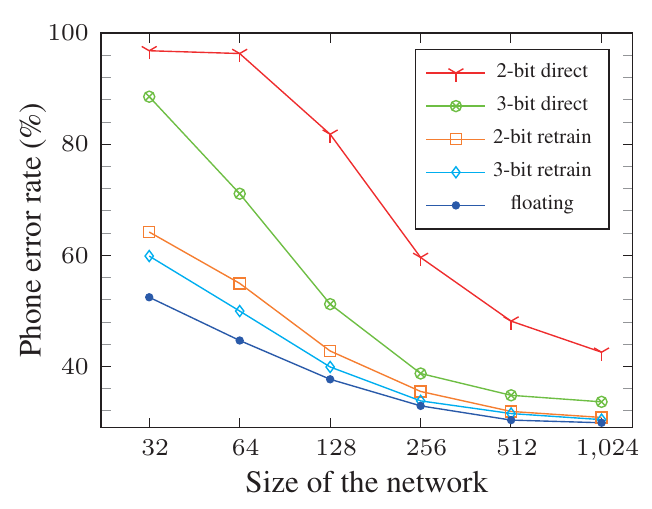}\label{fig_dnn_full_quant}
}
\subfloat[CNN]{
\includegraphics[width=0.45\linewidth]{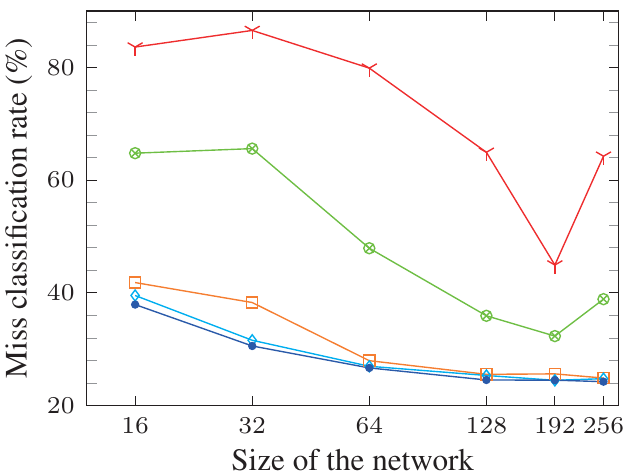}\label{fig_cnn_full_quant}
}
\caption{Comparison of retrain-based and direct quantization. All weights are quantized with ternary and 7-level weights. Note that each FCDNN has four hidden layers. In the figure (b), x-axis label ``16, 32, 64, 128, 192, 256'' represents the number of feature map is ``8-8-16, 16-16-32, 32-32-64, 64-64-128, 96-96-192, 128-128-256''.} \label{fig_full_quant}
\end{figure}

The fixed-point performance of the FCDNN is shown in \figurename~\ref{fig_dnn_full_quant}, where the number of hidden units in each layer varies. 
The performance of direct 2 bits (ternary levels), direct 3 bits (7-levels), retrain-based 2 bits, and retrain-based 3 bits are compared with the floating-point results.
We can find that the performance gap between the floating-point and the retrain-based fixed-point networks converges very fast as the network size grows. 
Direct quantization does not show good results at any network size. 
In this figure, the performance of the floating-point network almost saturates when the network size is about 1024. 
Note that the TIMIT corpus that is used for training has only 3 hours of data. 
Thus, the network with 1024 hidden units can be considered in the `\emph{training-data limited region}'. 
Here, the gap between the floating-point and the fixed-point networks almost vanishes when the network is in the `\emph{training-data limited region}'.
However, when the network size is limited, such as 32, 64, 128, or 256, there exists some performance gap between the floating-point and highly quantized networks even if retraining on the quantized networks is performed. 
The similar experiments are conducted for the CNN with varying feature map sizes, and the results are shown in \figurename~\ref{fig_cnn_full_quant}.
The configurations of the feature maps used for the experiments are 8-8-16, 16-16-32, 32-32-64, 64-64-128, 96-96-192, and 128-128-256. 
The size of the fully connected layer is not changed.
In this figure, the floating-point and the fixed-point performances also converge very fast as the number of feature maps increases. 
The floating-point performance saturates when the feature map size is 128-128-256, and the gap is less than 1\% when comparing the floating-point and the retrain-based 2-bit networks.
This suggests that a fairly high-performance feature extraction can be designed even using very low-precision weights if the number of feature maps can be increased. 
\section{Efficient DNN Design with Hardware Constraints }
\begin{figure}[h]
\centering
\subfloat[]{
\includegraphics[width=0.47\linewidth]{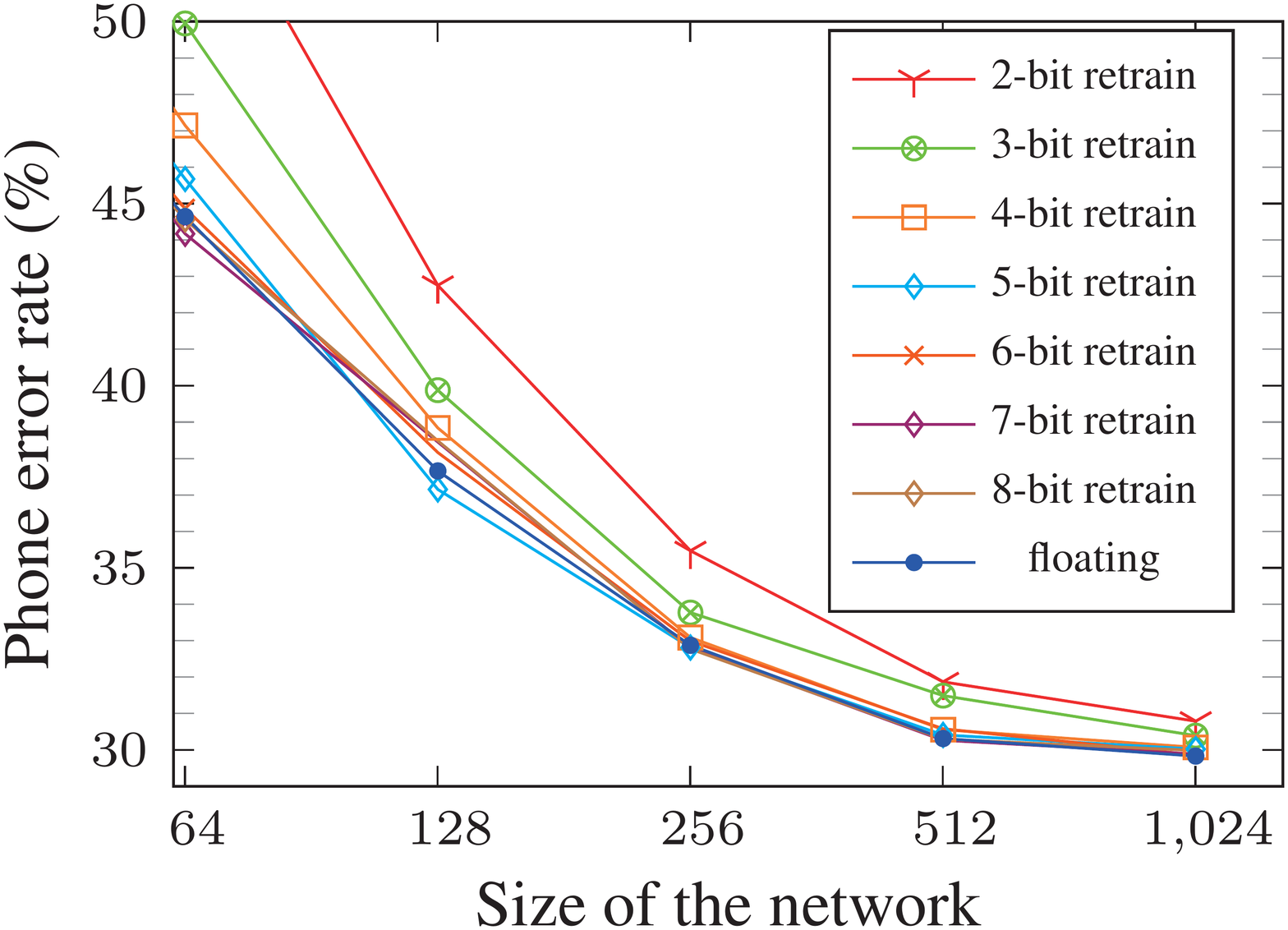}\label{fig_number_of_bits_b}
}
\subfloat[]{
\includegraphics[width=0.45\linewidth]{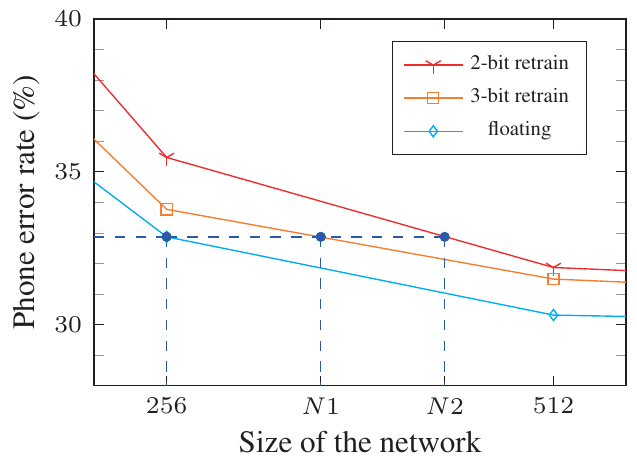}\label{fig_ecr_lines}
}
\caption{(a)  Framewise phone error rate of phoneme recognition FCDNNs with respect to the size of the networks for weights with quantization. (b) Obtaining an effective number of parameters for the uncompressed network.} \label{fig_number_of_bits}
\end{figure}

As the number of quantization levels decreases, the memory space needed is reduced at the cost of sacrificing the accuracy. Therefore, there can be a trade-off between the network size reduction and aggressive quantization. \figurename~\ref{fig_number_of_bits_b} shows the framewise phoneme error rate on TIMIT corpus while varying the layer size of FCDNNs with a various number of quantization bits from 2 to 8 bits. Note that the network has four hidden layers containing the same number of units. 

In this section, we propose a guideline for finding the optimal bit-widths when the desired accuracy or the network size is given. Note that we assume $2^n - 1$ quantization levels are represented by $n$ bits (i.e. 2 bits are required for representing a ternary weight). For simplicity, all layers are quantized with the same number of quantization levels.
%
%
%
\begin{figure}[h]
\centering
\includegraphics[width=0.95\linewidth]{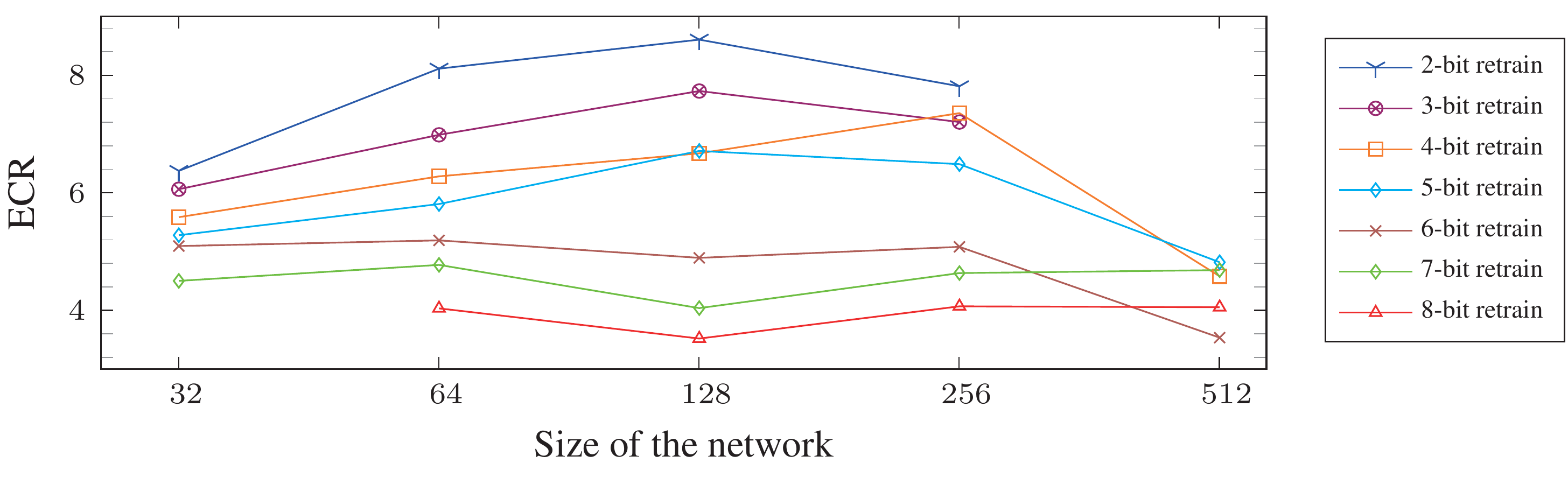}
\caption{Effective compression ratio (ECR) of fixed-point networks with respect to the layer size of reference floating-point networks.}\label{fig_ecr}
\end{figure}
Based on this observation, we introduce a metric called the \emph{effective compression ratio (ECR)}, which is defined as follows:
\begin{align}
ECR = \left[\frac{\;\text{Size of a floating-point network}}{{\text{Size of a quantized network showing the same performance}}}\right] \label{eq_ecr}
\end{align}
\figurename~\ref{fig_ecr_lines} describes how to compare the hardware efficiency of floating-point and fixed-point networks. In this figure, we assume the target performance of 32.87\% which can be obtained using a floating-point DNN with the network size of 256. This graph shows that the target performance can be obtained with the network size of $N1$ when 3 bits weights are used, and that of $N2$ when 2 bits quantization is employed. Thus, the compression obtained by 3-bit quantization can be roughly figured as $(32\times256^2)/(3\times N1^2)$. If $N1$ is very close to 256, there can be about 10 (= 32/3) times compression. But, if $N1$ is 512, the compression drops to only about 2.5 because the number of parameters of FCDNN is proportional to the square of the network size.   

The ECRs for various network sizes and quantization bits are shown in \figurename~\ref{fig_ecr}. The figure illustrates that the 2-bit or (maybe 1-bit) quantization can lead to the best compression ratio when the target performance is low, which means a high phoneme error rate in this example. However, for designing a fairly high-performance network, increasing the network size with severe quantization does not yield hardware efficient networks. The optimum number of bits for obtaining the performance corresponding to that of 512 sizes floating-point DNNs is 4 or 5 bits. Further reducing the word-length demands very increased size networks, and as a result, the total number of bits increases.    

\section{Conclusion}
\label{sec:conclusion}
Hardware efficient deep neural networks can be designed either by lowering the number of units in each layer or reducing the number of bits for weight quantization.  We evaluate the performance of fixed-point deep neural networks and analyze the trade-off between the complexity and the precision of the weights. This study shows that low-performance hardware efficient DNNs can be designed with severely quantized weights. In the low-performance region, the DNN performance increases very rapidly as the network size grows. Thus, it is possible to compensate for the quantization effects by slightly increasing the network size.  However, for a high-performance DNN design, compensation of quantization effects by increasing the network size is difficult, and thus severe quantization does not lead to efficient hardware design. The effective compression ratio is given for a DNN design when the network size and the precision vary.  

\subsubsection*{Acknowledgments}
This work was supported in part by the Brain Korea 21 Plus Project and the National Research Foundation of Korea (NRF) grant funded by the Korea government (MSIP) (No. 2015R1A2A1A10056051).


\bibliographystyle{nips_2016}
\bibliography{example_paper}

\end{document}